\documentclass[10pt]{article}
\usepackage[preprint]{tmlr}

\usepackage{amsmath,amsfonts,bm}

\def\eqref#1{equation~\ref{#1}}
\def\1{\bm{1}}

\DeclareMathAlphabet{\mathsfit}{\encodingdefault}{\sfdefault}{m}{sl}
\SetMathAlphabet{\mathsfit}{bold}{\encodingdefault}{\sfdefault}{bx}{n}

\usepackage{booktabs}
\usepackage{graphicx}
\usepackage[section]{placeins}
\usepackage{microtype}
\usepackage{hyperref}
\usepackage{url}

\hypersetup{
  hidelinks,
  pdftitle={LayerRAG-Bench: A Cross-Layer Reliability Benchmark for Agentic Retrieval-Augmented Generation},
  pdfauthor={Musa Shams},
  pdfsubject={Named preprint},
  pdfkeywords={retrieval-augmented generation, agentic RAG, reliability, benchmark, tool use}
}

\title{LayerRAG-Bench: A Cross-Layer Reliability Benchmark for Agentic Retrieval-Augmented Generation}
\author{\name Musa Shams\\
\addr Independent Researcher \quad \href{https://orcid.org/0009-0005-1015-5342}{ORCID: 0009-0005-1015-5342} \quad \href{https://github.com/MusaShams/layerrag-bench}{github.com/MusaShams/layerrag-bench}}

\begin{document}
\maketitle

\begin{abstract}
Agentic retrieval-augmented generation systems can produce answers that appear
grounded while failing at the evidence, tool-contract, authorization, or
session-state layer. We introduce LayerRAG-Bench, a controlled cross-layer
reliability benchmark with 8 enterprise domains, 240 tasks, 9 fault scenarios,
2 contract modes, and 38,880 live task-level records across nine models from
OpenAI, Anthropic, and Gemini. Schema normalization raises schema-drift success
from 0.000 to 0.913, but stale evidence, missing tool output, denied
permissions, and wrong-session context are not recovered by schema
normalization. Groundedness-only evaluation also produces substantial false
positives under stale and
wrong-session evidence. These results support a layer-specific evaluation
principle: a reliability intervention should be credited for repairing its
target layer without being mistaken for a universal fix.
\end{abstract}
\begingroup
\renewcommand{\thefootnote}{}
\footnotetext{Large language model tools were used to assist with editing,
code review, and manuscript preparation. The author is responsible for all
ideas, claims, experiments, analyses, and final text.}
\endgroup

\section{Introduction}
\label{sec:introduction}

Retrieval-augmented generation conditions model outputs on external evidence,
while agentic RAG adds tool calls, structured interfaces, permissions, and
persistent state~\citep{lewis2020rag,gao2023rag_survey,yao2023react}. Reliability
therefore depends on more than whether an answer is textually supported. A
response can faithfully summarize a superseded policy, cite evidence from the
wrong session, or proceed from incomplete tool output and still appear
grounded.

Existing retrieval and faithfulness metrics are necessary but can conflate
failures from different layers. This matters operationally because schema
normalization may repair a changed tool payload, whereas it cannot restore
missing evidence, override authorization, refresh a stale index, or correct
tenant context. Evaluation should distinguish those cases rather than collapse
them into one answer-quality score.

LayerRAG-Bench introduces controlled faults across retrieval, freshness,
metadata, tool contracts, permissions, completeness, and session state. It
combines deterministic regression tests with a live matrix of nine models,
nine scenarios, two contract modes, and 240 task rows per cell. The study makes
three contributions:
\begin{enumerate}
    \item a reproducible enterprise-style benchmark that maps nine scenarios to
    distinct operational layers;
    \item matched strict and repair conditions that test whether schema repair
    improves its intended layer without masking negative controls; and
    \item privacy-safe archives and hierarchical uncertainty analysis for
    recomputing all reported aggregate results.
\end{enumerate}

The main finding is structural rather than leaderboard-oriented. Contract
repair consistently recovers schema drift, while stale, unavailable,
unauthorized, incomplete, and wrong-session evidence requires separate
controls. LayerRAG-Bench is therefore complementary to retrieval, groundedness,
and general factuality benchmarks: it evaluates whether a mitigation fixes the
layer it claims to address.

\section{Benchmark and Methods}
\label{sec:methods}

\subsection{Benchmark design and threat model}

The \texttt{enterprise\_v2} corpus contains 8 enterprise domains, 80 policy
specifications, 160 documents, and 240 tasks. Each policy unit has a current
and superseded document plus three templated question variants. We treat these
variants as correlated coverage cases rather than 240 independent content
units. Documents expose policy prose and operational metadata; task-level
expected answers and required document identifiers remain evaluator-only.

LayerRAG-Bench targets non-adversarial operational faults inside a RAG stack:
stale indexing, schema drift, unavailable or partial tool output, conflicting
evidence, permission filtering, corrupted metadata, and wrong session state.
It does not claim to evaluate prompt injection, jailbreaks, data exfiltration,
or production access-control enforcement. A failure should not always be
``repaired'' by the model: denied evidence should trigger a safe non-answer,
and stale or wrong-session evidence should not receive credit merely because it
supports the generated text.

\begin{table}[t]
\centering
\caption{Fault scenarios, targeted reliability layers, and expected mitigation behavior.}
\label{tab:fault-layer-mapping}
\small
\begin{tabular}{p{0.24\textwidth}p{0.22\textwidth}p{0.46\textwidth}}
\toprule
Scenario & Targeted layer & Expected mitigation behavior \\
\midrule
clean & none & baseline should succeed \\
stale index & evidence freshness & requires fresh retrieval or index validation \\
schema drift & tool contract & contract repair should help \\
missing tool output & tool availability & requires retry, fallback, or abstention \\
conflicting evidence & evidence conflict & requires conflict detection and resolution \\
permission denied & access control & should not be bypassed by generation \\
partial tool timeout & tool completeness & requires timeout-aware abstention or recovery \\
corrupted metadata & metadata integrity & depends on whether content remains recoverable \\
wrong session state & session context & requires session isolation or context validation \\
\bottomrule
\end{tabular}
\end{table}

Table~\ref{tab:fault-layer-mapping} defines the intended failure layer and
mitigation boundary for each scenario. This mapping is the basis for the
paper's layer-specific interpretation.

\subsection{Evaluation protocol}

The deterministic harness evaluates all 240 tasks without model calls using
the reported \texttt{tfidf\_domain\_validity} retriever. It combines TF--IDF
matching with query-derived domain cues and status/year validity reranking; it
does not receive the task's labeled domain.

The live harness applies the same task definitions, retriever, scenarios,
contract modes, and scorer to nine models spanning OpenAI, Anthropic, and
Gemini. Crossing 9 models, 9 scenarios, 2 contract modes, and 240 task rows
produces 162 cells and 38,880 records. Strict mode rejects tool outputs that do
not match the expected schema. Repair mode applies bounded schema
normalization; it is not designed to reconstruct missing evidence, bypass
permissions, refresh an index, or repair session context.

A separate focused LangChain study compares uncontrolled retrieval and
generation with evidence-aware allow, abstain, reject, and bounded-retry
controls. It uses one model, eight tasks, nine evidence conditions, and two
arms, yielding 144 evaluated cells. Strict success treats abstention as
failure; safe resolution credits either a correct answer on answerable
conditions or a structured non-answer on unsafe conditions.

\subsection{Metrics}

For task \(t\), strict success is
\[
S_t = E_t \land R_t \land G_t \land V_t,
\]
where \(E_t\) is normalized exact match, \(R_t\) requires all task-specific
evidence identifiers to be retrieved, \(G_t\) requires at least one citation
and restricts citations to retrieved evidence, and \(V_t\) requires every tool
call to satisfy its schema. This conjunction is intentionally stricter than
answer correctness or groundedness alone. Abstentions remain strict failures
but are retained in diagnostic labels and, in the focused integration study,
the separate safe-resolution metric.

The historical live runner serialized a version-local answer field from
document metadata alongside answer-bearing prose. It did not expose task-level
expected answers or required document identifiers. Current code removes that
field. We therefore interpret historical live results as end-to-end pipeline
measurements, not as an isolated test of metadata-free reading comprehension.

\subsection{Artifacts}

Privacy-safe scored-record archives contain all 38,880 cross-provider records
and 144 focused-integration cells needed to recompute scenario aggregates,
repair gains, groundedness false positives, and bootstrap intervals without
provider calls. They omit raw model prose, free-form rationale, answer text,
document identifiers, latency, and cost. Full protocol settings and additional
audits appear in Appendix~\ref{app:protocol}.

\section{Related Work}
\label{sec:related-work}

Retrieval benchmarks such as DPR and BEIR evaluate relevance and ranking
quality~\citep{karpukhin2020dpr,thakur2021beir}. RAG evaluation extends this
toward answer correctness, citation quality, and faithfulness; RAGAS, ARES,
CRAG, and Self-RAG automate or adapt several of these dimensions
~\citep{lewis2020rag,gao2023rag_survey,es2023ragas,saadfalcon2023ares,yang2024crag,asai2023selfrag}.
These properties remain necessary, but evidence can be faithful and still be
superseded, unauthorized, incomplete, or associated with the wrong session.

Tool-use benchmarks study action selection, schema compliance, and structured
outputs~\citep{yao2023react,schick2023toolformer,qin2023toolbench}. Recent
diagnostic work localizes intermediate RAG and agent failures, while
reliability and enterprise benchmarks introduce production-like faults or
multidimensional evaluation
~\citep{lin2025ragcap,you2026agenticragtracer,jiao2026doctorrag,gupta2026reliabilitybench,sun2026enterpriseragbench,narita2026realworldrag}.

LayerRAG-Bench is complementary to these lines of work. Its distinct object is
a matched cross-layer fault matrix with negative controls: a mitigation is
credited when it repairs its intended operational layer without appearing to
solve unrelated freshness, authorization, completeness, or state failures.
This is consistent with broader systematizations of risks across retrieval,
memory, and tool-execution loops~\citep{mishra2026agenticragsok}.

\section{Results}
\label{sec:results}

\subsection{Deterministic benchmark behavior}

Table~\ref{tab:deterministic-enterprise-v2} reports the full deterministic
evaluation over 240 tasks and nine scenarios. It verifies that the benchmark
separates answerable conditions, contract failures, and evidence or state
failures before introducing model variability.

\begin{table}[t]
\centering
\caption{Deterministic enterprise v2 scenario results.}
\label{tab:deterministic-enterprise-v2}
\begin{tabular}{lrrr}
\toprule
Scenario & N & Success & Exact \\
\midrule
clean & 240 & 0.90 & 0.90 \\
stale\_index & 240 & 0.00 & 0.00 \\
schema\_drift & 240 & 0.00 & 0.00 \\
missing\_tool\_output & 240 & 0.00 & 0.00 \\
conflicting\_evidence & 240 & 0.90 & 0.90 \\
permission\_denied & 240 & 0.00 & 0.00 \\
partial\_tool\_timeout & 240 & 0.32 & 0.33 \\
corrupted\_metadata & 240 & 0.55 & 0.55 \\
wrong\_session\_state & 240 & 0.00 & 0.00 \\
\bottomrule
\end{tabular}
\end{table}

Clean and conflicting-evidence conditions are largely answerable, whereas
stale index, schema drift, missing output, permission denial, and wrong session
state fail for distinct reasons. In particular, stale and wrong-session
answers can remain grounded in retrieved documents while being operationally
incorrect.

\subsection{Retriever and context sensitivity}

A deterministic post-hoc audit evaluates the exact retriever interfaces on all
240 tasks without model calls. Table~\ref{tab:retriever-sensitivity} reports
required-document top-1 recall under clean, conflicting-evidence, and
corrupted-metadata conditions.

\begin{table}[!htbp]
\centering
\caption{Required-document top-1 recall under deterministic retriever
ablations. Each cell evaluates all 240 task rows. The explicit task-domain
filter is a sensitivity condition, not the reported retriever configuration.}
\label{tab:retriever-sensitivity}
\begin{tabular}{lrrr}
\toprule
Retriever condition & Clean & Conflicting evidence & Corrupted metadata \\
\midrule
Reported domain + validity & 0.896 & 0.508 & 0.554 \\
Domain only (validity removed) & 0.483 & 0.358 & 0.529 \\
Validity only (domain removed) & 0.975 & 0.558 & 0.554 \\
Plain TF--IDF & 0.529 & 0.392 & 0.529 \\
Explicit task-domain filter & 0.954 & 0.412 & 0.000 \\
\bottomrule
\end{tabular}

\end{table}

Validity awareness is the main source of current-version preference. Removing
validity reranking lowers clean top-1 recall from 0.896 to 0.483. Under
corrupted metadata, the reported retriever's required-document top-1 recall is
0.554 and its current-version preference rate is 0.562. The query-derived
domain heuristic is correct for 88 tasks,
incorrect for 14, and unresolved for 138.

One exact standalone query appears in both returns and legal tasks with
different answers. The reported retriever ranks the required document first
for only one of the two; explicit task-domain filtering resolves both.
However, explicit filtering is itself brittle to corrupted domain metadata,
where top-1 and top-3 recall both fall to zero. This establishes a central
boundary of the benchmark: query text alone may not identify the correct
domain or session, and contextual metadata is useful only when its integrity
is trustworthy.

\FloatBarrier

\subsection{Live cross-provider matrix}

The primary live result is the scenario--contract pattern in
Table~\ref{tab:cross-provider-full-matrix-scenario-contract}. Each row
aggregates 2,160 records: nine models times 240 task rows.

\begin{table}[!htbp]
\centering
\caption{Cross-provider success rates by scenario and contract mode.}
\label{tab:cross-provider-full-matrix-scenario-contract}
\begin{tabular}{llrr}
\toprule
Scenario & Contract & Records & Success \\
\midrule
Clean & Strict & 2,160 & 0.906 \\
Clean & Repair & 2,160 & 0.914 \\
Conflicting evidence & Strict & 2,160 & 0.904 \\
Conflicting evidence & Repair & 2,160 & 0.906 \\
Corrupted metadata & Strict & 2,160 & 0.900 \\
Corrupted metadata & Repair & 2,160 & 0.897 \\
Missing tool output & Strict & 2,160 & 0.000 \\
Missing tool output & Repair & 2,160 & 0.000 \\
Partial tool timeout & Strict & 2,160 & 0.314 \\
Partial tool timeout & Repair & 2,160 & 0.308 \\
Permission denied & Strict & 2,160 & 0.000 \\
Permission denied & Repair & 2,160 & 0.000 \\
Schema drift & Strict & 2,160 & 0.000 \\
Schema drift & Repair & 2,160 & 0.913 \\
Stale index & Strict & 2,160 & 0.000 \\
Stale index & Repair & 2,160 & 0.000 \\
Wrong session state & Strict & 2,160 & 0.000 \\
Wrong session state & Repair & 2,160 & 0.000 \\
\bottomrule
\end{tabular}

\end{table}

All 38,880 records have resolved binary success labels. Schema drift is the
dominant repairable fault: strict success is 0.000 and repair success is 0.913.
Stale index, missing tool output, permission denial, and wrong session state
remain at 0.000 under both modes. Partial timeout remains degraded under both
modes (0.314 strict; 0.308 repair).

The paired schema-drift improvement is 91.3 percentage points with a
hierarchical 95\% interval of 85.4--95.6 points. Its largest absolute
off-target contract change is 0.7 points, giving a specificity margin of
90.6 points.

\begin{table}[!htbp]
\centering
\caption{Groundedness false positives and paired strict-to-repair effects. GF
denotes records grounded in retrieved documents that nevertheless fail strict
benchmark success. Differences and intervals are percentage points.}
\label{tab:task-level-diagnostics}
\resizebox{\textwidth}{!}{
\begin{tabular}{lrrr}
\toprule
Scenario & GF strict & GF repair & Repair--strict $\Delta$ [95\% CI] \\
\midrule
Clean & 25 (1.2\%) & 22 (1.0\%) & +0.7 [-0.4, +2.3] \\
Conflicting evidence & 26 (1.2\%) & 21 (1.0\%) & +0.2 [-0.9, +1.7] \\
Corrupted metadata & 29 (1.3\%) & 29 (1.3\%) & -0.4 [-1.8, +1.0] \\
Missing tool output & 0 (0.0\%) & 0 (0.0\%) & +0.0 [+0.0, +0.0] \\
Partial tool timeout & 31 (1.4\%) & 32 (1.5\%) & -0.6 [-1.7, +0.1] \\
Permission denied & 0 (0.0\%) & 0 (0.0\%) & +0.0 [+0.0, +0.0] \\
Schema drift & 0 (0.0\%) & 22 (1.0\%) & +91.3 [+85.4, +95.6] \\
Stale index & 15 (0.7\%) & 9 (0.4\%) & +0.0 [+0.0, +0.0] \\
Wrong session state & 1676 (77.6\%) & 1703 (78.8\%) & +0.0 [+0.0, +0.0] \\
\bottomrule
\end{tabular}
}
\end{table}

Table~\ref{tab:task-level-diagnostics} shows why groundedness alone is
insufficient. Under wrong session state, 77.6\% of strict records and 78.8\% of
repair records are grounded yet unsuccessful; every grounded stale-index
record also fails strict success. Schema repair, by contrast, improves paired
task success by 91.3 points because the required evidence remains present and
only the contract shape is broken.

\subsection{Representative failure mechanisms}

Table~\ref{tab:representative-traces} contrasts one repairable contract fault
with evidence, authorization, and context failures.

\begin{table}[!htbp]
\centering
\caption{Representative task-level outcomes. ``Unknown'' denotes no usable
answer under the benchmark contract.}
\label{tab:representative-traces}
\small
\begin{tabular}{@{}p{0.17\textwidth}p{0.10\textwidth}p{0.21\textwidth}p{0.42\textwidth}@{}}
\toprule
Scenario & Mode & Expected / observed & Interpretation \\
\midrule
Schema drift & strict & 45 days / unknown &
The changed payload fails validation before evidence reaches answering. \\
Schema drift & repair & 45 days / 45 days &
Normalization restores the expected shape and the task succeeds. \\
Stale index & repair & 45 days / unknown &
The current policy is absent; schema repair cannot restore evidence. \\
Permission denied & repair & 45 days / unknown &
No authorized evidence is available, so the system should not guess. \\
Wrong session & repair & 45 days / 30 days &
The answer is grounded in text from the wrong session context. \\
\bottomrule
\end{tabular}
\end{table}

The paired schema-drift rows isolate the intended mechanism: repair restores a
changed representation of available evidence. The remaining cases require
freshness checks, access-aware abstention, timeout recovery, or session
isolation rather than schema normalization.

\subsection{Focused LangChain integration}

\begin{table}[!htbp]
\centering
\caption{Focused LangChain integration. Safe resolution credits correct
answers on answerable conditions and structured non-answers on unsafe ones.}
\label{tab:langchain-integration-results}
\begin{tabular}{lrr}
\toprule
Arm & Strict success & Safe resolution \\
\midrule
Baseline LangChain & 25/72 (34.7\%) & 36/72 (50.0\%) \\
LangChain + LayerRAG & 24/72 (33.3\%) & 72/72 (100.0\%) \\
\bottomrule
\end{tabular}
\end{table}

Within the LayerRAG arm, strict success is 24/24 on answerable conditions and
safe resolution is 48/48 on unsafe conditions. Clean-task strict success
remains 100\% in both arms, with no incorrect LayerRAG control interventions.
The study is intentionally narrow and supports safe handling, not a general
strict-accuracy claim.

\subsection{Interpretation}

The evidence supports three conclusions. First, contract repair should be
evaluated against contract faults: it recovers schema drift but does not restore
missing evidence or state. Second, groundedness alone is insufficient when
freshness, authorization, completeness, and context matter. Third, similar
failure shapes across providers indicate that stronger base models do not
eliminate the need for layer-specific system controls. Provider-level and
repair-gain breakdowns appear in Appendix~\ref{app:live-details}.

\section{Interpreting Fixed-Budget Results}
\label{sec:uncertainty}

The live matrix is a fixed-budget systems evaluation, not a powered commercial
model ranking. Each model contributes 4,320 evaluations, each provider 12,960,
and each scenario--contract row 2,160. The strongest claims concern repeated
failure structure rather than small provider differences.

Strict and repair outcomes are paired by provider, model, scenario, and task.
The repair-minus-strict interval uses a hierarchical bootstrap that resamples
provider--model units and then task pairs within each selected unit, preserving
the nested design. Schema drift improves by 0.913 with a 95\% interval of
[0.854, 0.956]. Clean, conflicting-evidence, corrupted-metadata, and
partial-timeout intervals include zero; missing output, permission denial,
stale index, and wrong session effects are exactly zero under both modes.

These results support a mechanism-level interpretation. Schema drift is
repairable because the needed evidence remains present and the failure is
localized to the tool contract. The negative controls remain unresolved because
their missing conditions are not schema-format problems.

\section{Reproducibility}
\label{sec:reproducibility}

The public repository includes benchmark construction, deterministic
evaluation, table export, tests, and aggregate-recomputation code. Fixed
specifications generate the corpus, tasks, and scenarios. Live execution uses
explicit call budgets, while deterministic workflows and public-result
recomputation require no provider access.

The committed privacy-safe archives reproduce scenario and provider
aggregates, repair gains, groundedness false-positive counts, hierarchical
bootstrap intervals, and focused LangChain results. Prompt hashes are present
for 28,080 of 38,880 records; the remaining records terminate before prompt
construction because retrieval is unavailable or invalid. The released
archives exclude credentials, raw model prose, unrestricted traces, and
non-allowlisted identifying metadata.

Appendix~\ref{app:protocol} reports provider settings, archive boundaries, and
historical-run caveats. The manuscript source and public artifacts are
versioned together in the repository linked on the first page.

\section{Limitations and Future Work}
\label{sec:limitations}

The corpus is synthetic and policy-like. This enables controlled faults but
does not reproduce the ambiguity, document diversity, user behavior, or
organizational complexity of production deployments. The 240 task rows are
three variants of 80 policy units, so linguistic variants are correlated rather
than independent samples.

The reported retriever infers domain from query wording; it was unresolved on
138 tasks and incorrect on 14 in the deterministic audit. Historical live
prompts also included a version-local answer field in document metadata,
although answer-bearing prose contained the same value and task-level labels
were not exposed. Current code removes the field. The historical cross-provider
and focused-integration results should therefore be interpreted as measurements
of the pre-hardening end-to-end pipeline, not metadata-free reading
comprehension.

The cross-provider matrix is broad in task count but limited to nine evaluated
models and a fixed budget. The focused LangChain study is narrower still: one
model, eight selected tasks, and one completion per cell. Neither study
establishes provider rankings, production latency advantages, or general
security guarantees.

Future extensions should evaluate freshness-aware retrieval, session
isolation, timeout recovery, policy-based abstention, human escalation,
multi-turn and multi-hop tasks, hierarchical permissions, adversarial document
conflicts, open-source models, and repeated temporal replications. The current
contribution is the layer-specific evaluation framework, not exhaustive
coverage of every RAG reliability or security failure.

\section{Conclusion}
\label{sec:conclusion}

LayerRAG-Bench treats agentic RAG reliability as a layered systems problem. In
38,880 cross-provider records, schema repair raises schema-drift success from
0.000 to 0.913 but does not repair stale evidence, missing output, denied
permissions, partial timeouts, or wrong-session context. Groundedness can
therefore be necessary while remaining operationally insufficient.

The broader principle is that mitigations should be evaluated against the
layer they are designed to repair. Retrieval improvements, schema
normalization, freshness validation, access-aware abstention, timeout recovery,
and session isolation address different mechanisms. LayerRAG-Bench makes those
boundaries measurable and provides reproducible negative controls against
overstating a narrow intervention as a universal reliability solution.

\FloatBarrier
\clearpage
\phantomsection
\label{tmlr:references-start}
\bibliographystyle{tmlr}
\bibliography{references}

@inproceedings{lewis2020rag,
  title = {Retrieval-Augmented Generation for Knowledge-Intensive {NLP} Tasks},
  author = {Lewis, Patrick and Perez, Ethan and Piktus, Aleksandra and Petroni, Fabio and Karpukhin, Vladimir and Goyal, Naman and K{\"u}ttler, Heinrich and Lewis, Mike and Yih, Wen-tau and Rockt{\"a}schel, Tim and Riedel, Sebastian and Kiela, Douwe},
  booktitle = {Advances in Neural Information Processing Systems},
  year = {2020}
}

@inproceedings{karpukhin2020dpr,
  title = {Dense Passage Retrieval for Open-Domain Question Answering},
  author = {Karpukhin, Vladimir and Oguz, Barlas and Min, Sewon and Lewis, Patrick and Wu, Ledell and Edunov, Sergey and Chen, Danqi and Yih, Wen-tau},
  booktitle = {Proceedings of the 2020 Conference on Empirical Methods in Natural Language Processing},
  year = {2020}
}

@inproceedings{thakur2021beir,
  title = {{BEIR}: A Heterogeneous Benchmark for Zero-shot Evaluation of Information Retrieval Models},
  author = {Thakur, Nandan and Reimers, Nils and R{\"u}ckl{\'e}, Andreas and Srivastava, Abhishek and Gurevych, Iryna},
  booktitle = {Proceedings of the Neural Information Processing Systems Track on Datasets and Benchmarks},
  year = {2021}
}

@article{gao2023rag_survey,
  title = {Retrieval-Augmented Generation for Large Language Models: A Survey},
  author = {Gao, Yunfan and Xiong, Yun and Gao, Xinyu and Jia, Kangxiang and Pan, Jinliu and Bi, Yuxi and Dai, Yi and Sun, Jiawei and Wang, Haofen},
  journal = {arXiv preprint arXiv:2312.10997},
  year = {2023}
}

@inproceedings{yao2023react,
  title = {{ReAct}: Synergizing Reasoning and Acting in Language Models},
  author = {Yao, Shunyu and Zhao, Jeffrey and Yu, Dian and Du, Nan and Shafran, Izhak and Narasimhan, Karthik and Cao, Yuan},
  booktitle = {International Conference on Learning Representations},
  year = {2023}
}

@inproceedings{schick2023toolformer,
  title = {{Toolformer}: Language Models Can Teach Themselves to Use Tools},
  author = {Schick, Timo and Dwivedi-Yu, Jane and Dess{\`i}, Roberto and Raileanu, Roberta and Lomeli, Maria and Hambro, Eric and Zettlemoyer, Luke and Cancedda, Nicola and Scialom, Thomas},
  booktitle = {Advances in Neural Information Processing Systems},
  year = {2023}
}

@article{qin2023toolbench,
  title = {{ToolBench}: Towards Tool-Augmented Large Language Models},
  author = {Qin, Yujia and Cai, Shihao and Wang, Xiaozhi and Chen, Yining and Liu, Zhouhan and Liang, Yankai and Liu, Zhi and Han, Xiaojie and Han, Xu and Liu, Zhiyuan and Sun, Maosong and Zhou, Jie},
  journal = {arXiv preprint arXiv:2307.16789},
  year = {2023}
}

@article{es2023ragas,
  title={{RAGAS}: Automated Evaluation of Retrieval Augmented Generation},
  author={Es, Shahul and James, Jithin and Espinosa-Anke, Luis and Schockaert, Steven},
  journal={arXiv preprint arXiv:2309.15217},
  year={2023}
}

@article{saadfalcon2023ares,
  title={{ARES}: An Automated Evaluation Framework for Retrieval-Augmented Generation Systems},
  author={Saad-Falcon, Jon and Khattab, Omar and Potts, Christopher and Zaharia, Matei},
  journal={arXiv preprint arXiv:2311.09476},
  year={2023}
}

@article{yang2024crag,
  title={{CRAG} -- Comprehensive {RAG} Benchmark},
  author={Yang, Xiao and Sun, Kai and Xin, Hao and Sun, Yushi and Bhalla, Nikita and Chen, Xiangsen and Choudhary, Sajal and Gui, Rongze Daniel and Jiang, Ziran Will and Jiang, Ziyu and Kong, Lingkun and Moran, Brian and Wang, Jiaqi and Xu, Yifan Ethan and Yan, An and Yang, Chenyu and Yuan, Eting and Zha, Hanwen and Tang, Nan and Chen, Lei and Scheffer, Nicolas and Liu, Yue and Shah, Nirav and Wanga, Rakesh and Kumar, Anuj and Yih, Wen-tau and Dong, Xin Luna},
  journal={arXiv preprint arXiv:2406.04744},
  year={2024}
}

@article{asai2023selfrag,
  title={{Self-RAG}: Learning to Retrieve, Generate, and Critique through Self-Reflection},
  author={Asai, Akari and Wu, Zeqiu and Wang, Yizhong and Sil, Avirup and Hajishirzi, Hannaneh},
  journal={arXiv preprint arXiv:2310.11511},
  year={2023}
}

@article{gupta2026reliabilitybench,
  title={{ReliabilityBench}: Evaluating {LLM} Agent Reliability Under Production-Like Stress Conditions},
  author={Gupta, Aayush},
  journal={arXiv preprint arXiv:2601.06112},
  year={2026}
}

@article{lin2025ragcap,
  title={{RAGCap-Bench}: Benchmarking Capabilities of {LLM}s in Agentic Retrieval Augmented Generation Systems},
  author={Lin, Jingru and Zhang, Chen and Liu, Stephen Y. and Li, Haizhou},
  journal={arXiv preprint arXiv:2510.13910},
  year={2025}
}

@article{you2026agenticragtracer,
  title={{AgenticRAGTracer}: A Hop-Aware Benchmark for Diagnosing Multi-Step Retrieval Reasoning in Agentic {RAG}},
  author={You, Qijie and Yu, Wenkai and Zhang, Wentao},
  journal={arXiv preprint arXiv:2602.19127},
  year={2026}
}

@article{jiao2026doctorrag,
  title={{Doctor-RAG}: A Failure-Aware Repair Framework for Agentic Retrieval-Augmented Generation},
  author={Jiao, Shuguang and Huang, Chengkai and Qi, Shuhan and Wang, Xuan and Li, Yifan and Weng, Quanchi and Liu, Lingchuan and Cai, Xunliang and Yao, Lina},
  journal={arXiv preprint arXiv:2604.00865},
  year={2026}
}

@article{narita2026realworldrag,
  title={Overcoming the ``Impracticality'' of {RAG}: Proposing a Real-World Benchmark and Multi-Dimensional Diagnostic Framework},
  author={Narita, Kenichirou and Peng, Siqi and Fukui, Taku and Yamada, Moyuru and Munakata, Satoshi and Takahashi, Satoru},
  journal={arXiv preprint arXiv:2604.02640},
  year={2026}
}

@article{sun2026enterpriseragbench,
  title={{EnterpriseRAG-Bench}: A {RAG} Benchmark for Company Internal Knowledge},
  author={Sun, Yuhong and Rahmfeld, Joachim and Weaver, Chris and Chen, Weijia and Desai, Roshan and Huang, Wenxi and Butler, Mark H.},
  journal={arXiv preprint arXiv:2605.05253},
  year={2026}
}

@article{mishra2026agenticragsok,
  title={{SoK}: Agentic Retrieval-Augmented Generation ({RAG}): Taxonomy, Architectures, Evaluation, and Research Directions},
  author={Mishra, Saroj and Niroula, Suman and Yadav, Umesh and Thakur, Dilip and Gyawali, Srijan and Gaire, Shiva},
  journal={arXiv preprint arXiv:2603.07379},
  year={2026}
}

\clearpage
\appendix
\section{Additional Live-Matrix Results}
\label{app:live-details}

\begin{table}[!htbp]
\centering
\caption{Largest strict-to-repair gains.}
\label{tab:cross-provider-full-matrix-repair-gains}
\resizebox{\textwidth}{!}{
\begin{tabular}{lllrrr}
\toprule
Provider & Model & Scenario & Strict & Repair & Delta \\
\midrule
Anthropic & claude-sonnet-5 & Schema drift & 0.000 & 0.958 & 0.958 \\
OpenAI & gpt-5.5 & Schema drift & 0.000 & 0.958 & 0.958 \\
OpenAI & gpt-5.4 & Schema drift & 0.000 & 0.954 & 0.954 \\
OpenAI & gpt-5.4-mini & Schema drift & 0.000 & 0.954 & 0.954 \\
Anthropic & claude-haiku-4-5-20251001 & Schema drift & 0.000 & 0.950 & 0.950 \\
Gemini & gemini-3.1-flash-lite & Schema drift & 0.000 & 0.950 & 0.950 \\
Anthropic & claude-opus-4-8 & Schema drift & 0.000 & 0.946 & 0.946 \\
Gemini & gemini-3.5-flash & Schema drift & 0.000 & 0.825 & 0.825 \\
Gemini & gemini-3.1-pro-preview & Schema drift & 0.000 & 0.721 & 0.721 \\
Gemini & gemini-3.5-flash & Clean & 0.792 & 0.829 & 0.038 \\
Gemini & gemini-3.1-pro-preview & Conflicting evidence & 0.683 & 0.713 & 0.029 \\
Gemini & gemini-3.1-pro-preview & Clean & 0.708 & 0.729 & 0.021 \\
OpenAI & gpt-5.4-mini & Clean & 0.950 & 0.958 & 0.008 \\
Anthropic & claude-sonnet-5 & Conflicting evidence & 0.954 & 0.958 & 0.004 \\
Gemini & gemini-3.5-flash & Conflicting evidence & 0.758 & 0.762 & 0.004 \\
\bottomrule
\end{tabular}
}
\end{table}

\begin{table}[!htbp]
\centering
\caption{Provider-level averages, reported as supporting context rather than
commercial-model rankings.}
\label{tab:cross-provider-full-matrix-provider}
\begin{tabular}{lrr}
\toprule
Provider & Records & Success \\
\midrule
Anthropic & 12,960 & 0.407 \\
Gemini & 12,960 & 0.343 \\
OpenAI & 12,960 & 0.410 \\
\bottomrule
\end{tabular}

\end{table}

Provider averages combine clean tasks, repairable contract failures, and
negative controls that should remain unresolved. They are therefore not
ordinary capability scores. The central repeated pattern is the scenario-level
specificity of schema repair.

\section{Protocol and Artifact Details}
\label{app:protocol}

The eight benchmark domains are returns, paid time off, support, security,
finance, human resources, legal, and IT operations. The reported retriever uses
\texttt{top\_k=3}. Provider clients use a 180-second request timeout, a
600-token output cap, three total attempts, and a 2-second linear retry
backoff. Request temperature was not explicitly supplied, so endpoints used
their API defaults; Anthropic requests use API version
\texttt{2023-06-01}.

The reported matrix contains 54 model--scenario--contract cells for each of
OpenAI, Anthropic, and Gemini (162 total). Run manifests prevent double
counting of mirrored outputs. The privacy-safe
archives contain fields needed for aggregate recomputation but omit raw model
output, answer text, free-form rationale, document identifiers, latency, cost,
and non-allowlisted metadata.

The benchmark does not treat permission denial, stale evidence, or wrong
session state as faults that a model should answer through. Those scenarios
test whether system boundaries remain visible. Prompt injection, malicious
documents, exfiltration, and production authorization certification are outside
the present threat model.

\end{document}